\documentclass[english]{article}
\usepackage[T1]{fontenc}
\usepackage[latin9]{inputenc}
\usepackage{fancyhdr}
\usepackage{color}
\usepackage{babel}
\usepackage{float}
\usepackage[numbers,square]{natbib}
\usepackage[unicode=true,pdfusetitle,
 bookmarks=true,bookmarksnumbered=false,bookmarksopen=false,
 breaklinks=true,pdfborder={0 0 0},pdfborderstyle={},backref=false,colorlinks=true]
 {hyperref}
\usepackage{hyperref}
\usepackage{xurl}
\usepackage{breakurl}
\hypersetup{breaklinks=true}

\makeatletter

\providecommand{\tabularnewline}{\\}

\@ifundefined{date}{}{\date{}}

\usepackage{graphicx}
\usepackage{titlesec}
\usepackage{url}

\renewcommand{\thesection}{\Alph{section}}

\titleformat{\section}[hang]{\bfseries\Large}{\thesection.}{0.4em}{}
\titleformat{\subsection}[hang]{\bfseries\large}{\thesubsection.}{0.4em}{}
\titleformat{\subsubsection}[hang]{\bfseries}{\thesubsubsection.}{0.4em}{}

\title{\textbf{}}
\author{Michael Dalvean \\[1ex]\\ School of Computer Science and Software Engineering\\ University of Western Australia\\ michael.dalvean@research.uwa.edu.au
\textit{} \\
\textit{}}
\date{}

\usepackage{array}
\newcolumntype{L}[1]{>{\raggedright\let\newline\\\arraybackslash\hspace{0pt}}m{#1}}
\newcolumntype{C}[1]{>{\centering\let\newline\\\arraybackslash\hspace{0pt}}m{#1}}

\makeatother

\begin{document}
\pagebreak{}
\title{A Stochastic Analysis of the Linguistic Provenance of English Place
Names}
\maketitle
\begin{abstract}
In English place name analysis, meanings are often derived from the
resemblance of roots in place names to topographical features, proper
names and/or habitation terms in one of the languages that have had
an influence on English place names. The problem here is that it is
sometimes difficult to determine the base language to use to interpret
the roots. The purpose of this paper is to stochastically determine
the resemblance between 18799 English place names and 84687 place
names from Ireland, Scotland, Wales, Denmark, Norway, Sweden, France,
Germany, the Netherlands and Ancient Rome. Each English place name
is ranked according to the extent to which it resembles place names
from the other countries, and this provides a basis for determining
the likely language to use to interpret the place name. A number of
observations can be made using the ranking provided. In particular,
it is found that `Harlington' is the most archetypically English place
name in the English sample, and `Anna' is the least. Furthermore,
it is found that the place names in the non-English datasets are most
similar to Norwegian place names and least similar to Welsh place
names. 
\end{abstract}
%\textit{Keywords: place name, linguistics, machine learning} \\
Word count: 9011 \\

\lfoot{{\footnotesize{}English Place Names}}

\section{Introduction}

English place names are often analysed through the lenses of Germanic,
Celtic and Latin languages, and the starting point for analysis is
what the place name may mean in a given language. Highly technical
linguistic, etymological and philological methods are used to determine
the meaning of place names by technical experts with a knowledge of
the history and traditions underlying place name conventions. The
problem is that different but equally plausible explanations for the
same place name can be derived using these methods. The differences
often come from different starting points in terms of the underlying
language used for the analysis. In many cases, there seems to be no
objective way to determine which language to use to analyse place
names.

The purpose of this paper is to use objective methods to derive the
language origins of English place names. The approach is to use computational
linguistics to determine the influence of ten European languages on
English place names. In this way we can identify the probable language
to use to begin the etymological exercise or, alternatively, eliminate
languages on the basis of probable non-influence. 

Without written records we may never know the actual original meaning
of words. However, we can determine the linguistic origin of place
names with a high degree of probability. By using the place names
of other countries, we can determine the extent to which the morphology
of a given English place name resembles the morphology of place names
from other countries. In this way, we can determine the likely linguistic
origin of individual place names. 

The method entails analysing English place names by comparing them
to the place names of ten other major European countries that have
had a high level of linguistic and cultural influence on England and
the place names of England. These countries are France, Germany, Ancient
Rome, Norway, Sweden, Denmark, the Netherlands, Scotland, Ireland
and Wales. The idea is to rank each individual English place name
in terms of the extent to which it resembles the place names of England
as opposed to the place names of the other ten European countries. 

By distilling the languages of place names, we get an objective probability
that a particular language is the most likely source of the place
names. By assigning probabilities to several languages in this way,
we can partial out the relative contributions of each language on
a given place name. In the case of English place names, there are
several candidate languages that can enter into the provenance of
a place name. A traditional explanation of the name is often based
on a local topographical feature and the resemblance of the topographical
feature in a given language to the place name. The etymology of `London',
for example, has been linked to its location on the Thames and words
connoting water. Gigot, for example, links the `Burgundo-Gothic' root
`lohna', meaning `puddle' or `pond' with `London' on the basis that
this root connotes the Thames \citep{gigot1974notes} cited in \citep{coates1998new}.
The problem with this approach is that elements of a place name can
be linked with numerous topographical words in other languages. Thus,
`Londonthorpe' in Lincolnshire is described as being derived from
the Old Scandinavian lundr {[}grove{]} + thorp {[}outlying settlement{]}
\citep{mills2011dictionary}. Thus, we have two suggestions for the
origin of `London' from two different languages using two different
topographical features. The selection of the association between a
topographical term in a given language and a topographical feature
associated with the relevant place name is a rational basis upon which
to determine the meaning of a place name. However, if we do not know
the linguistic provenance of a place name, we will not know what language
with which to begin the analysis. 

The approach taken in this paper is that the first step in place name
analysis is to determine the linguistic provenance of place names.
Linguistic provenance in this study is determined by looking at the
similarity of the place names of England to the place names of countries
where the languages spoken are likely to have influenced the formation
and/or the development of English place names. For a given place name
and for a given comparison country, we generate a score based on the
extent to which the English place name resembles the place names from
the other country. If the score indicates that the place name is more
like other English place names than like the place names of the other
country, then we can conclude that the other language did not have
a significant influence on that place name. However, if the score
indicates that the place name is like the place names of the other
country, we can conclude that the country did have an influence on
the place name and we would direct our etymological investigation
accordingly. 

It should be stressed that the search for the meaning of place names
is not the purpose of this paper. This paper merely provides a means
for determining a starting point for the search for meaning in that
it provides a likely language basis for a place name. 

One of the benefits of determining the likely language origins of
English place names is that it enables us to rank English place names
in terms of their English or `Anglo-Saxon' content, and this enables
us to identify the `archetypal' English place name. We will see that
`Harlington' is the most archetypal English place name, while `Anna'
is the least. We will also be able to determine the relative linguistic
difference between English place names and the place names of other
countries. In particular, we will see that the difference between
English place names and Scottish place names is the smallest in the
cohort of ten languages, whereas the difference between English and
Ancient Roman place names is the largest. Finally, we will see that
there are 475 English place names that defy classification as English.

Please note that there are some differences in the terminology used
in computational linguistics and formal linguistics. The terminology
of computational linguistics is used in this paper.

\section{Previous Work}

The standard analysis of English place names seeks to derive an understanding
of English culture and settlement patters based on the interpretation
of the meaning of place names \citep{scott2011directions}. The method
\begin{quotation}
...focuses largely on empirical, at times empiricist, enquiry, working
through extant historical records in order to chart the history of
name spellings, and infer name origins from established philological
paradigms, drawing where possible on comparative material from related
disciplines such as history and archaeology. \citep[p29]{scott2011directions}
\end{quotation}
There is a great deal of emphasis on the linguistic provenance of
place names in the analysis because the interpretation of the meaning
of the place name will depend on the language basis of the place name.
For example, Coates \citep{coates1998new} which looks at the possible
roots of the place name `London' together with explanations as to
how phonological influences over time in Britain may have affected
the morphology of the name. Coates holds that `London' is based on
a river name `plowonidon', which is made up of two roots from Old
European, an early Indo-European language now largely known only from
hydronyms. 

There is another vein of place name research in which the provenance
of the place name is not at issue, although the meaning is. An example
is Hough \citep{hough1996place}, which examines the use of Scottish
place names as an aid in the interpretation of Old English legal history.
Similarly, O'Grady \citep{o2014judicial} uses place names and archaeology
to examine the nature of court locations in medieval Scotland. Cuthbertson
\citep{cuthbertson2018use} uses Scottish place names to examine Scots
lexicography. For example, the topographical evidence from Scottish
place names is that `doors` meant `passage` in Scots. 

In yet another vein of place name research, neither the provenance
nor the meaning are at issue, but the patterns of types of place names
are used to shed light on other cultural and historical phenomena.
For example, Fekete \citep{fekete2015anglo} analyses the Scandinavian
and Anglo-Saxon elements of English place names and finds that in
the Danelaw area in the north-east of England, there is evidence of
`code-mixing' such that many place names have both Scandinavian and
Anglo-Saxon elements. Fekete's study is an excellent example of pattern
analysis in that the computational process involves looking for specific
patterns such as the Germanic habitation elements `-ton', `-by', and
`-thorpe'. 

A recent innovation in place name research has been the use of machine
learning. In this approach, instead of looking for specific patterns,
the machine learning algorithm trawls through the data in an attempt
to derive patterns, and the analysis proceeds using the patterns derived
thereby. Much of the previous machine learning work on place names
has been associated with information retrieval and database applications.
Marchardo \textit{et al} \citep{machado2011ontological}, for example,
use computational methods to extract geographical names and terms
from web documents. Here, the issue is that many words can be names
of entities other than places, and an algorithm designed to extract
place names must be trained to identify the word as denoting a place
name. Similarly, Berragan \textit{et al} \citep{berragan2022transformer}
use computational methods to extract geographical place names from
geographical texts. Hu \textit{et al} \citep{hu2022gazpne} use rules,
gazetteers and deep leaning to extract place names from micro blogs.
A study that combines text analysis with machine learning is Bensalem
\citep{bensalem2010toponym}, which details an algorithm for disambiguating
identical place names by determining the likely reference using contextual
data in the text from which it is drawn. 

The current study brings together aspects of the standard approach
to place name analysis as well as machine learning aspects. The focus
of the current study is broadly similar to the standard approach to
place name analysis in that it seeks to determine the language origin
of place names. In contrast to the standard approach, however, the
current study uses machine learning to determine the provenance of
place names. It is beyond the scope of the current paper to attempt
to analyse the place name using methods such as those used by Gigot
\citep{gigot1974notes}, Mills \citep{mills2011dictionary} and Coates
\citep{coates1998new}. The application of etymological, linguistic
and philological expertise is the next step after determining the
language provenance of the place name. 

\section{A Stochastic Approach to Place Name Analysis}

The current study takes a stochastic approach to place name analysis.
It seeks to determine the language basis of English place names, but
not the meaning. 

The most significant difference in the approach of the current study
and that of the more traditional approaches is that the current study
uses the place names of other countries as benchmarks for `English'
morphological structure. The structure of the place names of England
is compared with the structure of the place names of ten other European
countries. The commonalities in the English place names are derived
from this comparison. For example, the element `-ton' comes from Anglo
Saxon `-tun', meaning `enclosed space. It is the most common element
in English place names. The algorithm in this study does not specifically
look for the element `-ton', which is the approach of the more standard
rule-based `molecular' approaches to place name analysis. Instead,
the current study compares the `atomic' letter placement of the sample
of English place names with the letter placement of the place names
of other countries. It transpires that, in comparison to the other
place names, English place names have a structure such that where
the last letter = `n', the second last letter = `o', and the third
last letter = `t', there is a high probability that the place name
is English. Importantly, each of these variables is independent of
the other. Thus, in the element `-tan', the presence of the `t' in
the third last place and the `n' in the last place has the same effect
on the algorithm's assessment of the place name as the `t' and the
`n' in a place name ending in `ton'. However, the presence of the
`a' in the `-tan' element has a different effect in comparison to
the `o' in `-ton'. The different weightings for the presence/absence
of particular letters in particular positions in the place names are
derived by comparison with the place names of England to the place
names of other countries. As we will see below, nine of the ten most
`archetypal' Anglo-Saxon place names have the element `-ton', and
the fact that this element was identified as an important indicator
of Anglo-Saxon origin without any rule-based search indicates the
efficacy of the stochastic `atomic' approach. 

A benefit of this approach is that the training algorithm can search
for patterns that are not known in advance. The deterministic rule-based
search for patterns is efficient, but it is not as efficient as a
system that can both derive the same set of rules that the researcher
knows as well as detect rules that the researcher does not know. It
is this process inherent in the machine learning methodology that
the current study benefits from.

The focus on individual letter placement as a method of breaking down
place names into the variables used in the analysis is derived from
`Benford's Law' \citep{nigrini1997use}. This law applies to the distribution
of single digit numerals in particular places of multi digit numerals.
For a given distribution of multi-digit numerals, the rate of occurrence
of the numeral `1' in the first position will be 0.301, that of the
numeral `2' will be 0.176, with a decreasing rate up to the digit
`9' which will occur in the first place of multi-digit numerals at
a rate of 0.041. The law states that the probability of a digit `d'
being in first place in a sample of multi-digit numerals is
\[
log_{10}(1+1/d).
\]

This concept was applied to letters in words in Yan \textit{et al}
\citep{yan2018benford}, in which it was found that the most common
first letter in the words of nine works of English literature by English
and American authors' novels was `t', which occurred at an average
rate of 16.26\%. The second most common first letter was `a', which
occurred at a rate of 11.24\%. It was found in that study that when
the rates of occurrence of each of the 26 letters in the first letter
of words were ranked from highest to lowest, the decay followed a
similar declining rate to numerals in Benford's law, with `x' having
the lowest rate of occurrence in first place.

The current study extends the idea of the importance of the position
of letters by creating a set of variables for place names based on
letter placement. The idea here is that it is likely that there will
be differences in the structure of place names of different countries.
We will see below, for example, that there is a significant difference
in the place names of Ancient Rome as opposed to England in that the
former are highly statistically significantly likely to end in `a'
or `i' in comparison to the latter. Furthermore, this study extends
the analysis to letters in the second, third and fourth letters as
well as the last, second last, third last and fourth last letters. 

Thus, the first set of variables consists of 26 binary variables where
the value = 1 if the fist letter is a...z, and = 0 if the letter is
not present. The second, third and fourth sets of variables are similar
but are based on the second, third and fourth letter of the place
name respectively. Thus, there are four sets of binary variables based
on the letters in the first four positions in place names = 4 {*}
26 = 104. A similar set of variables is created for the four last
letters of place names. 

Another six binary variables were created indicating the presence/absence
anywhere in the place name of individual vowels (A,E,I,O U and Y).
Six rate variables were created by dividing each of these vowel binaries
by the length of the name. One variable was created by taking the
total number of vowels in each name and dividing by the length of
the name. Twenty variables were created by creating binaries for the
presence/absence anywhere in the name of each of the 20 consonants,
and another 20 variables were created by dividing each of these by
the length of the name. Name length and entropy (bits per letter)
were included in the variables. 

In this way, each place name is broken into 263 variables, most of
which are based on individual letters and letter placement.

Before we continue, it is worthwhile addressing an issue that may
be considered as problematic in the methodology: place names have
changed since their original inception, and therefore using the modern
versions of place names will lead to invalid results. This objection
can be addressed by considering the stochastic nature of the analysis.
The modern name `Birkby' has undergone several changes since it was
first recorded in 1086 as `Bretebi' \citep{carroll2020identifying}.
The first letter has remained, and the `bi' has transformed into `by'.
The `r' has moved one place forwards and the vowel `e' has transformed
into `i'. The `t' has become `k'. The `e' before the `by' element
has gone, thereby shortening the name by one letter. The ratio of
vowels (A,E,I,O,U and Y) to consonants has fallen from 3:4 to 2:4.
Let us first consider the `by' element in the modern name. In the
sample of 18799 English place names used in this analysis, there are
no place names ending in `bi', and there are 391 that end with `by'.
We know that the `by' element is a Scandinavian settlement term. There
is clearly a pattern in language change in that the `bi' has gone
from English names and has been replaced with `by'. It might be suggested
that the `bi' could have transformed in some cases to `be'. However,
inspection of the data shows that 183 place names end in `be', 182
of which are due to the element `-combe', which has a different etymology.
In short, there has been a systematic change in the form of `bi' to
by. However, we cannot be as definite about some of the other changes
that have occurred. Consider the change of `t' to `k' in the middle
of the original name. Has this shift been systematic? Without further
analysis, we will not know. However, let us assume that the change
has not been systematic and that in half of the English place names
with `t' in the middle of the name the `t' has changed to `k' while
the remaining half have stayed the same. The algorithm will detect
a pattern for English place names such that a `t' or a `k' in the
4th position of the name is associated with English place names. Thus,
whether the change is systematic or partial, the pattern is likely
to be detected because there are binary variables indicating the presence/absence
of individual letters in specific positions in the place name. Similar
arguments could be used to explain why the other changes in `Bretebi'
are not likely to result in the inability of the algorithm to identify
`Birkby' as `English'. The vowel and consonant variables, as well
as the length and entropy variables are included so that even if the
changes in language and orthographic conventions are not systematic,
there is the possibility of detecting the relevant signals in the
data. We will address this issue empirically below when we examine
the algorithm's ability to distinguish between place names of Old
Norse origin and Anglo-Saxon Origin. We will see that despite the
fact that modern spellings are used to train the algorithm, the algorithm
is able to distinguish between the two classes. 

\section{Data}

\subsection{Data Sources}

Where possible, the place names used in this study were accessed from
the official repositories of place name data in the respective countries.
However, in some cases this was not possible, so alternative sources
were used. Sources for the initial place names were as follow: 
\noindent \begin{flushleft}
-England, Scotland, Wales: Index of Place Names in Great Britain (September
2019) \citep{english}
\par\end{flushleft}

\noindent \begin{flushleft}
-France: Liste des communes au 1er janvier 2022 \citep{french}
\par\end{flushleft}

\noindent \begin{flushleft}
-Germany: Gemeinden in Deutschland nach Fläche, Bevölkerung und Postleitzahl
am 30.09.2021 \citep{german}
\par\end{flushleft}

\noindent \begin{flushleft}
-Denmark Autoriserede stednavne i Danmark \citep{denmark} 
\par\end{flushleft}

\noindent \begin{flushleft}
-Sweden, Norway, Ireland: Country Files (GNS) \citep{denswenorire}
\par\end{flushleft}

\noindent \begin{flushleft}
-Ancient Rome: Pleiades Ancient Names Database \citep{rome}
\par\end{flushleft}

The US GNS data was used for Sweden, Ireland and Norway because, although
individual place name data is available online for these countries,
it was not possible to download the data in toto. In the case of the
Danish data, only entries labeled as `By' = Town, and `Bydel' = District
were included because many of the names were for features of the landscape
such as tunnels and ports rather than place names.

\subsection{Data Processing and Cleaning}

Only single names were used in this study. Therefore all names with
more than one word, such as `West London', or with hyphens (`Aix-en-Provence')
were removed. The reasoning here is that the simple structure is more
likely to carry a fundamental signal than a compound structure.

All letters were transformed into the most close approximation to
the 26 Latin characters of the English alphabet. The reason for this
is that there are certain characters that would provide a signal that
could not be applied to all names. The umlaut, for example, occurs
predominantly in the Germanic samples and therefore would provide a biased
signal. Letters with diacritics were transformed to their closest Latin
characters using Python's unidecode function. Thus, characters
such as `ø' and `æ' were changed to `o' and `ae' respectively. The
German `ß' was changed to `ss' and the French `œ' was changed to `oe'. 

A search for repetitions of names within each country was made and
repeated names were removed. Each country file therefore represents
a list of unique names for the country.

A second removal process involved removing names that were repeated
across countries. The reason for this is that if the same name occurs
in two countries, there is no ground truth as to its most likely origin. 

Sample sizes are reported in Table 1.

\begin{table}
\caption{Place Name Sample Sizes}

\centering{}%
\begin{tabular}{|l|l|}
\hline 
Country & Sample Size\tabularnewline
\hline 
\hline 
England & 18799\tabularnewline
\hline 
Denmark & 5493\tabularnewline
\hline 
Norway & 5056\tabularnewline
\hline 
Sweden & 20794\tabularnewline
\hline 
Ireland & 9928\tabularnewline
\hline 
Scotland & 4203\tabularnewline
\hline 
Wales & 2814\tabularnewline
\hline 
Ancient Rome & 1688\tabularnewline
\hline 
Germany & 8699\tabularnewline
\hline 
France & 18175\tabularnewline
\hline 
Netherlands & 7837\tabularnewline
\hline 
\end{tabular}
\end{table}

\section{Modeling}

In initial experiments it was found that the most accurate classifiers
were one-to-one binary classifiers for each of the England-Other pairs.
Thus, ten binary classifiers were created using the England sample
as the target. 

Ten datasets were created using the English names as the `cases' (binary
= 1), and each of the other name samples as `controls' (binary = 0).
This led to significantly imbalanced datasets. The case of the English-Roman
data set was the most extreme, with more than ten times the number
of English names than Ancient Roman names. The solution for this problem
was to use a variation on Synthetic Minority Oversampling Technique
(SMOTE). The procedure was implemented using the Python imbalanced-learn
library. The basic SMOTE methodology is to synthesise cases in the
minority sample (`oversampling') using linear interpolation to increase
the minority class. The SMOTE-ENN (Synthetic Minority Oversampling
Technique-Edited Nearest Neighbor) methodology extends this by removing
from both classes any case that is misclassified by the majority class
of its three nearest neighbours \citep{batista2004study}. This method
has been shown to be efficient in modeling imbalanced datasets. It
should be stressed that the SMOTE-ENN procedure are only applied to
the training procedure: the test results were generated using the
held out folds at each of the ten stages of the 10-fold procedure.
Scores for each name are those generated where the name was in the
held out fold. The test for the efficacy of the SMOTE-ENN procedure is
the classification accuracy of the held out samples. As we will see
below, the classification accuracy for all classifiers is high.

The induction method chosen was random forest classifier. This algorithm
has been found to work well with the SMOTE-ENN procedure \citep{hemmat2016sla}. The implementation in the current study uses the random forest from Python's scikit-learn library with n\_estimators=100, max\_depth=None,
min\_samples\_split=5. Accuracy metrics (cut point = .5) for each
classifier are reported in the results section below. 

\section{Results}

\subsection{Accuracy}

Table 2 shows the accuracy metrics of the classifiers. The average
accuracy of the classifiers is 92\%. The lowest score is for Scotland
(83\%) and the highest for Ancient Rome (98\%). This level of accuracy
indicates that the classifiers are all able to efficiently distinguish
between the English and non-English place names using the variables
derived from the place names.

\begin{table}[H]
	\caption{Classifier Accuracies}
	
	\centering{}%
	\begin{tabular}{|l|c|c|c|}
		\hline 
		& Accuracy & Sensitivity & Specificity\tabularnewline
		\hline 
		\hline 
		England-Denmark & 0.94 & 0.94 & 0.92\tabularnewline
		\hline 
		England-Sweden & 0.94 & 0.92 & 0.96\tabularnewline
		\hline 
		England-Norway & 0.94 & 0.94 & 0.94\tabularnewline
		\hline 
		England-Ireland & 0.90 & 0.92 & 0.85\tabularnewline
		\hline 
		England-Scotland & 0.82 & 0.86 & 0.65\tabularnewline
		\hline 
		England-Wales & 0.92 & 0.96 & 0.66\tabularnewline
		\hline 
		England-France & 0.91 & 0.88 & 0.93\tabularnewline
		\hline 
		England-Rome & 0.98 & 0.99 & 0.88\tabularnewline
		\hline 
		England-Germany & 0.92 & 0.93 & 0.90\tabularnewline
		\hline 
		England-Netherlands & 0.91 & 0.92 & 0.91\tabularnewline
		\hline 
		Average: & 0.92 & 0.93 & 0.86\tabularnewline
		\hline 
	\end{tabular}
\end{table}

Inspection of the data indicates that the relatively low accuracy
for Scotland is due to the low specificity of 65\%. The specificity
is low because there is a large number of place names that are `English'
in style and form. This is also the case for Wales, which explains
the low specificity of 67\%.

A further test of accuracy is to see the rate correct classification
of English place names by the classifiers. The starting point for
this analysis is to calculate the average score for each place name
across the ten classifiers. For example, the place name `Laira' scores
the following on each classifier: Denmark - 0.56, Norway - 0.13, Sweden
- 0.09, France - 0.05, Germany - 0.38, Netherlands - 0.15, Rome -
0.05, Ireland - 0.10, Scotland - 0.03, Wales - 0.45. The average of
all these scores is 0.20. Using the standard machine learning cut
point of 0.5, we would classify this name as non-English as it scores
below 0.5. Applying the `ensemble of classifiers' to all 18799 English
place names in the same way, the accuracy is 97\%. However, we will
see that some of the misclassified names are misclassified because
they have a highly non-English structure. Thus, it is not necessarily
that the classifiers cannot detect the English-Other content of the
misclassified names, but that some of the names themselves are not
`English'. 

\subsection{Accuracy Test Using Old Norse and Old English Place Names}

In order to confirm the ability of the method to correctly classify
place names, a test was conducted using samples of place names deriving
from Old English and Old Norse/Scandinavian extracted from the Key
to English Place-Names at the Institute for Name-Studies at the University
of Nottingham \citep{nottingham_names}. To generate the Old English
sample, a search for names based on the following languages was made:
Old English, Anglian, Kentish, Mercian, Northumbrian (incl. Old N),
West-Saxon. This yielded 12507 names (including duplicates). A search
was made in the Derivation column for the term `Old English', and
those names that did not have the term were excluded, reducing the
sample to 12203. Any name with the text `Norse' or `Scand' in the
Derivation column was removed, giving a sample of 11301. Names of
more than one word, hyphenated names and names for which multiple
versions were given, as designated by a forward slash were removed,
and removal of duplicates resulted in final sample of 6485. 

The Old Norse/Scandinavian sample was generated by doing a search
for place names derived from the following languages: Old Norse, Old
Danish, Old East Scandinavian, Old Norwegian, Old West Scandinavian.
This generated a sample of 1719 place names. A search for `English',
`Saxon', `Mercian', `Northumbrian', `Kentish' and `Anglian' was made
in the Derivation column and, if present, the place name was removed.
This reduced the sample to 694. Finally, names of more than one word,
hyphenated names and names for which multiple versions were given,
as designated by a forward slash, were removed, yielding a final sample
of 541. Removal of duplicates resulted in a sample of 494. Average
scores on the Scandinavian classifiers for the Old English and Old
Norse samples are shown in Table 3.

\begin{table}[H]

\caption{OE and ON Place Name Scores on Scandinavian and English Classifiers}

\centering{}%
\begin{tabular*}{1\textwidth}{@{\extracolsep{\fill}}|c|c|c|c|c|c|}
\hline 
 & {\footnotesize{}Eng-Den} & {\footnotesize{}Eng-Swe} & {\footnotesize{}Eng-Nor} & {\footnotesize{}Ave' Scand'} & {\footnotesize{}Eng-Other}\tabularnewline
\hline 
\hline 
{\footnotesize{}OE Av (n=6485)} & {\footnotesize{}0.907 } & {\footnotesize{}0.921} & {\footnotesize{}0.892} & {\footnotesize{}0.906} & {\footnotesize{}0.884}\tabularnewline
\hline 
{\footnotesize{}ON A(n=494)} & {\footnotesize{}0.799} & {\footnotesize{}0.879} & {\footnotesize{}0.752} & {\footnotesize{}0.810} & {\footnotesize{}0.851}\tabularnewline
\hline 
{\footnotesize{}T-test} & {\footnotesize{}<.001} & {\footnotesize{}<.001} & {\footnotesize{}<.001} & {\footnotesize{}<.001} & {\footnotesize{}<.001}\tabularnewline
\hline 
{\footnotesize{}Mann-Whitney} & {\footnotesize{}<.001} & {\footnotesize{}<.001} & {\footnotesize{}<.001} & {\footnotesize{}<.001} & {\footnotesize{}<.001}\tabularnewline
\hline 
\end{tabular*}
\end{table}

Scores for the three Scandinavian classifiers (Eng-Den, Eng-Swe and
Eng-Nor) are averaged to give a Scandinavian average. The Eng-Other
score shows the scores for the ensemble (average) of ten classifiers,
including the three Scandinavian classifiers. T-tests (two tailed)
were conducted using Welch's test for the Eng-Den, Eng-Swe, Eng-Nor
and Ave' Scand data because the OE and ON data are from independent
samples and the n and variances are unequal. Results for the Mann-Whitney
test are also provided as the data are not normally distributed. For
the Eng-Other OE and ON data, an independent two tailed t-test was
done because the variances are equal. As expected, the scores for
the ON place names are significantly (p<.001) lower than the scores
for the OE place names on all three Scandinavian classifiers as well
as the Eng-Other ensemble on both the T-test and the Mann-Whitney
test. This indicates that the classifiers are efficient in distinguishing
between place names of ON and OE origin. This is an interesting finding
given that the classifiers were trained using modern English and Scandinavian
place names. 

\section{Discussion}

The classification accuracy for the ensemble of classifiers of English
place names of 97\% (cutpoint = .5) indicates that the ensemble classifier
is efficient. Furthermore, all ten individual classifiers are efficient,
as indicated by the average accuracy of 92\%. These results mean that the
actual scores generated for each place name are likely to be a good
indication of the English-Other content of the place name. The higher
this score, the higher the English content. We can therefore rank
each place name by the extent to which it represents an English (Anglo-Saxon)
place name as opposed to a place name with content from one or more
languages from the other ten countries. Table 4 shows the ten top scoring English place names. For
comparison, the ten lowest scoring English place names are also shown.

\begin{table}[H]
	\caption{Ten Highest/Lowest Scoring Place Names}
	
	\centering{}%
	\begin{tabular}{|l|c|}
		\hline 
		Place name & England-Other Score\tabularnewline
		\hline 
		\hline 
		Harlington & 0.999\tabularnewline
		\hline 
		Widdington & 0.998\tabularnewline
		\hline 
		Colworth  & 0.998\tabularnewline
		\hline 
		Beckington & 0.998\tabularnewline
		\hline 
		Didlington & 0.998\tabularnewline
		\hline 
		Toddington & 0.998\tabularnewline
		\hline 
		Lowthorpe & 0.998\tabularnewline
		\hline 
		Tiddington & 0.998\tabularnewline
		\hline 
		Bedlington & 0.998\tabularnewline
		\hline 
		Ridlington & 0.998\tabularnewline
		\hline 
		\hline 
		Danum & 0.231\tabularnewline
		\hline 
		Menna & 0.228\tabularnewline
		\hline 
		Moira & 0.227\tabularnewline
		\hline 
		Arun & 0.224\tabularnewline
		\hline 
		Belgravia & 0.223\tabularnewline
		\hline 
		Lamanva & 0.218\tabularnewline
		\hline 
		Laira & 0.200\tabularnewline
		\hline 
		Alma & 0.179\tabularnewline
		\hline 
		Lana & 0.166\tabularnewline
		\hline 
		Anna & 0.157\tabularnewline
		\hline 
	\end{tabular}
\end{table}

The place name with the highest
English content is `Harlington', which scores 0.999 on the English-Other
scale. The place name with the lowest English-Other score is `Anna',
which scores 0.157 . Table 5 breaks down the scores for these place
names in relation to all ten classifiers as well as the ensemble English-Other
score.

\begin{table}[H]
\caption{Scores for `Harlington' and `Anna'}

\begin{centering}
\begin{tabular}{|l|c|c|}
\hline 
 & Harlington & Anna\tabularnewline
\hline 
\hline 
Classifier & Score & Score\tabularnewline
\hline 
England-Denmark & 1.000 & 0.436\tabularnewline
\hline 
England-Sweden & 0.994 & 0.001\tabularnewline
\hline 
England-Norway & 1.000 & 0.038\tabularnewline
\hline 
England-France & 1.000 & 0.043\tabularnewline
\hline 
England-Germany & 1.000 & 0.238\tabularnewline
\hline 
England-Netherlands & 1.000 & 0.311\tabularnewline
\hline 
England-Ancient Rome & 1.00 & 0.047\tabularnewline
\hline 
England-Ireland & 1.000 & 0.061\tabularnewline
\hline 
England-Scotland & 0.998 & 0.148\tabularnewline
\hline 
England-Wales & 1.000 & 0.251\tabularnewline
\hline 
Average (England-Other) & 0.999 & 0.157\tabularnewline
\hline 
\end{tabular}
\par\end{centering}
\end{table}

It should be noted that the scores for all ten top scoring English names are effectively equal to 1 on the English-Other score. The ranking is based on five decimal places, but only three are shown. Thus, these are all archetypal `English' names in relation to the ten comparison datasets.

Of the `non-English' names,`Anna', scores lowest, having a 15.7\% probability of being English. If we look at the
individual classifier scores, we can see that Anna's lowest score
is on the England-Sweden classifier. The score of 0.001 indicates that
`Anna' has no statistical resemblance to the general sample of English
place names but, instead, has a complete statistical resemblance to general
Swedish place names. The highest resemblance of `Anna' to an English
name occurs with the England-Denmark classifier. The score of 0.436
indicates that it has a 43.6\% probability of being English in comparison
to Danish place names.

The ensemble score is useful in itself to identify the English content
in place names. However, the individual classifiers can be used to
individually distil the underlying contribution of each of the
other ten languages entering into the etymology of English place names.
If we rank the place names by score on the classifier of a given country,
we get a ranking of the relative resemblance of each English place
name in relation to the place names of the given country. For example,
if we rank the place names on the basis of the England-Denmark classifier,
the English place name with the highest Danish resemblance (lowest
England-Denmark resemblance) is Sedrup, which scores 0.01 on the England-Denmark classifier. At the other end of the scale, there are 2749 English
place names which score 1 on the England-Denmark classifier, including eight of the ten archetypal English place names in Table 4, the exceptions being Toddington and Bedlington which score 0.990 and 0.993 respectively. 

The foregoing analysis shows that it is possible to distil the `English'
content from place names. The implication of this is that most place
names are composites of linguistic components from cultures that have
had input into place names in England. This is not a controversial
proposition; the discussion of place names is characterised by historical
accounts of how place names have been created from various linguistic
influences. However, the above analysis is unique in that it provides
a method of measuring the input of various languages in the creation
of place names without specifically identifying features to search
for, such as the elements `-by', `-ton' and `-dale'. The benefit of
this is that the classifier can find both structures that are known
and those that are not. 

In the following section we will see how having a fine-grained place
name classifier can be used to address a number of issues in the place
name research.

\subsection{Resemblance of English to Other Place Names}

The scores for place names on each of the ten individual classifiers
give us an insight into which of the ten countries' place names the
English place names are most similar to. Table 6 shows the average
scores for the 18799 English place names for each of the classifiers
in the `Mean Score' column ordered by ascending accuracy. 

\begin{table}[H]
\caption{Mean Scores of English Place Names on Ten Classifiers}

\begin{centering}
\begin{tabular}{|l|c|}
\hline 
Classifier & Mean Score\tabularnewline
\hline 
\hline 
England-Scotland & 0.737\tabularnewline
\hline 
England-Ireland & 0.810\tabularnewline
\hline 
England-France & 0.815\tabularnewline
\hline 
England-Netherlands & 0.837\tabularnewline
\hline 
England-Germany & 0.844\tabularnewline
\hline 
England-Sweden & 0.848\tabularnewline
\hline 
England-Wales & 0.850\tabularnewline
\hline 
England-Denmark & 0.877\tabularnewline
\hline 
England-Norway & 0.889\tabularnewline
\hline 
England-Rome & 0.949\tabularnewline
\hline 
\end{tabular}
\par\end{centering}
\end{table}

If we consider the Mean Score for the England-Scotland classifier,
we can see that the average score for English place names using this
classifier is 0.737. At the other end of the spectrum, on the last
line of the table, the average score on the England-Rome classifier
is 0.949. What this indicates is that the England-Scotland classifier
is less efficient than the England-Rome classifier. The reason for
this is that there is a greater similarity between Scottish and English
place names than there is between Ancient Roman and English place
names. The classifier trained on Scottish and English place names
has a more difficult task distinguishing between English and Scottish
names because they are more similar, whereas the classifier trained
on the English and Roman place names is more efficient because there
is a great difference between the Roman and English place names.

What we can determine from Table 6 above is that there is a declining
similarity between English place names and those of the other ten
countries as we go down the table. The place names of countries used
to train classifiers in the top of the table are more similar to English
place names than those used to train classifiers in the lower segment
of the table. In short, the classifier accuracies are proxies for
place name similarity. This gives us a good insight into the influences
on English place names. For example, we can tell that the similarity
of English place names to those of Scotland, Ireland, France, the
Netherlands and Germany is greater than the similarity to any of the
Scandinavian place names. This is an interesting observation given
the significant inflow of Scandinavian settlers in England during
the period in which Anglo-Saxon place names were being applied to
English localities. However, the direction of causation may be difficult
to establish. Consider the case of Scotland: Scots is the Scottish
descendant of Old-English, and as such we might expect that there
would be significant similarities between Scottish and English place
names. Therefore, we cannot make the claim that Scottish influenced
English place names. Similarly, the resemblance of English place names
to Irish place names may be due to the adoption of English names in
Irish localities due to the political, military and economic influence
of England on Ireland. In the case of France, the Netherlands and
Germany, there is a good historical basis for claiming that the direction
of causation is from these continental countries to England due to
known historical migration patterns. Importantly, these migration
patterns were permanent, in that the settlers remained in England,
unlike the situation with Roman settlers who returned to Rome in the
early fifth century. This may explain the relative lack of influence
of Roman place names on English place names despite four centuries
of occupation. Finally, the low level of similarity between English
and Welsh place names is an interesting observation given the close
proximity of England and Wales. We can speculate that the reason for
this is that there was no systematic Anglo-Saxon settlement in Wales
after the departure of the Romans from Wales in the fifth century,
and the subsequent political influence likely to affect Welsh place
names came from the the French-speaking Norman conquerors of England
who turned their attention to Wales in the latter decades of the eleventh
century. Thus, there was little opportunity for Anglo-Saxon place
name conventions to be applied to Wales. 

\subsection{Broader Language Influences}

So far we have focused on English place names. However, the scores
on the ten country classifiers enable us to make some observations
about the broader language influences in Europe. We have scores for
English place names from ten classifiers. Each vector of 18799 scores
represents the similarity between the place names of an individual
country and the place names of England. If we look at the correlation
between the scores generated by the classifiers of two countries,
we get a measure of how similar the naming conventions are between
the two countries. That is, even though the scores are for English
place names, the differences between scores generated by different
classifiers are due to the interpretation of each classifier of the
English place name, and these differences are due to the classifiers
being trained on the place names of their own country. Correlations
between the scores of countries shows how similar their own place
names are to each other. 

Table 7 shows the inter-correlations between the scores for all ten
classifiers. The highest score in each column is highlighted in bold.
Note that all correlations are significant at the p<.01 level.

\begin{table}[H]
	\small
	
	\caption{Inter-correlations between classifier scores}
	
	\begin{tabular}{|c|c|c|c|c|c|c|c|c|c|c|}
	\hline 
	& Den & Nor & Swe & Ire & Scot & Wal & Fra & Ger & Net & Rom\tabularnewline
	\hline 
	\hline 
Den &  & 0.76 & 0.75 & 0.32 & 0.37 & 0.18 & 0.45 & 0.46 & 0.61 & 0.34\tabularnewline
\hline 
Nor & \textbf{0.76} &  & \textbf{0.84} & 0.39 & 0.43 & 0.33 & 0.5 & 0.53 & 0.69 & 0.47\tabularnewline
\hline 
Swe & 0.75 & \textbf{0.84} &  & 0.44 & 0.44 & 0.31 & 0.48 & 0.44 & 0.56 & 0.46\tabularnewline
\hline 
Ire & 0.32 & 0.39 & 0.44 &  & \textbf{0.62} & \textbf{0.48} & 0.51 & 0.35 & 0.37 & 0.43\tabularnewline
\hline 
Scot & 0.37 & 0.43 & 0.44 & \textbf{0.62} &  & 0.42 & 0.46 & 0.46 & 0.42 & 0.45\tabularnewline
\hline 
Wal & 0.18 & 0.33 & 0.31 & 0.48 & 0.42 &  & 0.46 & 0.37 & 0.32 & 0.37\tabularnewline
\hline 
Fra & 0.45 & 0.5 & 0.48 & 0.51 & 0.46 & 0.46 &  & 0.41 & 0.48 & \textbf{0.65}\tabularnewline
\hline 
Ger & 0.46 & 0.53 & 0.44 & 0.35 & 0.46 & 0.37 & 0.41 &  & \textbf{0.75} & 0.36\tabularnewline
\hline 
Net & 0.61 & 0.69 & 0.56 & 0.37 & 0.42 & 0.32 & 0.48 & \textbf{0.75} &  & 0.42\tabularnewline
\hline 
Rom & 0.34 & 0.47 & 0.46 & 0.43 & 0.45 & 0.37 & \textbf{0.65} & 0.36 & 0.42 & \tabularnewline
\hline 
Mean & 0.47 & \textbf{0.55} & 0.52 & 0.43 & 0.45 & 0.36 & 0.49 & 0.46 & 0.51 & 0.44\tabularnewline
\hline 
	\end{tabular}
	
\end{table}

In the first column, we can see that the Danish scores are most highly
correlated with the Norwegian scores (0.76). The Norwegian and Swedish columns show that these two sets of scores are the most highly correlated at 0.84. These correlations are uncontroversial given the linguistic
similarities among the Scandinavian countries. Similarly, the Irish and Scottish scores are most highly correlated with each other at 0.62, most likely due to their Gaelic origin. Interestingly, the Welsh scores have a relatively low correlation with the other two Gaelic languages at 0.48 for Ireland and 0.42 for Scotland.  French scores are most highly correlated with the Roman scores
(0.65). Again, given the Latin origin of French, this is uncontroversial.
The German scores are most highly correlated with the Netherlands
scores (0.75), due to both linguistic and geographical proximity.

The last row of the table gives the average score for each column.
This shows for each classifier the average correlation for that classifier
with the scores of the other 9 countries. The highest average correlation
is 0.55 for the Norwegian classifier. This shows that the scores generated
by the Norwegian classifier are the most highly correlated with the
scores of the other classifiers. In other words, Norwegian place name
conventions are more similar to the naming conventions of the other
countries than the place name conventions of any other individual
country. The Welsh naming conventions are clearly the least similar,
with an average correlation with the scores of other countries of
0.36. Again, these observations are uncontroversial given the historical
perambulatory inclinations of the Norwegians, and the relative non-perambulatory
inclinations of the Welsh. The Norwegians (Vikings) spent approximately
600 years venturing from their home to other areas of Europe, and
it is likely that, as in England, they left their linguistic mark
in the place names of the areas they visited or settled. This is in
contrast to the Welsh who have tended to be highly centralised in
Wales. 

One controversial observation is the relative lack of influence of
Ancient Roman naming conventions on the countries represented. The
average correlation of the Ancient Roman scores to the scores of the
other classifiers is 0.44. The Romans conquered the land mass of what
is now France, which explains the high correlation of Roman and French
classifier scores. However, the Romans only conquered the western
edge of what is now Germany, thereby leaving the bulk of `Germania'
relatively uninfluenced by their place naming conventions (r = 0.36).
Despite Ancient Roman presence in Wales and southern Scotland, there
was a relatively small influence of Ancient Rome on Welsh and Scottish
place names, with correlations of 0.37 and 0.45 respectively. Thus,
their influence on place names was greatest in France. Despite the
lack of physical conquest, the Romans had a high level of cultural
influence on all of Europe. Therefore, it is interesting to note that
their influence on the place names of Europe was less than seven of
the other countries represented. This may be an indication that in
the case of place names, military and political hegemony are more
important than cultural hegemony.

\section{Future Work and Conclusions}

The intention of this paper has been to identify the most likely linguistic
influences on English place names. By comparing each English place
name with the place names of ten western European countries, we are
able to apply a score to each English place name indicating whether
it is more like other English place names or more like the place names
of other countries. By using a large sample of place names from both
England and Europe, it is possible to determine with a high degree
of probability whether an English place name has the morphological
structure of an English place name or that from one of the other countries.

The use of the place names of other countries to partial out the influence
on English place names of those other countries is a novel procedure.
The number of place names in each sample is such that the results,
albeit stochastic, have a high degree of validity.

Another novel aspect is the method of breaking down each place name
into 263 mainly `atomic' variables. We have seen that this method
has been able to find patterns in the data that more `molecular' or
`rule-based' procedures are able to find, as well as finding others
that were not previously known. For example, the greater the number
of vowels divided by the length of a place name, the more likely it
is to be a Roman, as opposed to an English place name. 

The vast majority of English place names (97\%) are more like English
place names than those of other countries. However, there are English
place names that are `more English' than others. We have seen that
`Harlington' has more similarity to other English place names than
any other English place name. We have also seen that there are some
place names that are unrelated to English place names. 

The relative classifier accuracies were used as a proxy for the difference
between English place names and the place names of other countries.
The classifier accuracy for Ancient Rome was the highest, indicating
that Ancient Roman place names are the most different from English
place names, while the accuracy for the Scottish classifier was the
lowest, indicating that the Scottish place names are the most similar
to English place names. 

Finally, the analysis enabled us to look at some broad influences
on place name patterns. We saw that the Norwegian names are the most
similar to the names of the other 9 countries, while the Welsh names
are the least. We also saw that Ancient Roman names have a relatively
minor resemblance to the place names of the other countries given
the historical influence of Ancient Rome.

There are a number of ways the foregoing analysis could be extended. 

One extension would be to see how the geographical distribution of
more `English' place names differs from the geographical distribution
of less `English' place names. We have scores that can provide a ranking
method for English place names, and a subsequent study could be done
indicating the settlement patterns of the various different language
groups. Work has been done on settlement patterns in the Danelaw region
of the UK, showing a preponderance of Scandinavian names in areas
such as Cleveland, Yorkshire, Lindsay and Leicestershire \citep{stenton1942presidential}.
It would be interesting to see what the place name settlement patterns
are for the English place names that score low (have a high `other'
content) on each of the ten classifiers. We would most likely confirm
the aforementioned study on settlement patterns in the Danelaw areas,
with the likelihood that place names scoring low on the England-Denmark
classifier most likely being centered around the Danelaw areas. Similarly,
we might expect those place names scoring low in the England-Scotland
and England-Wales classifiers to be clustered around the Scottish
and Welsh borders respectively. However, it would be interesting to
see if there is a particular pattern in relation to the 471 place
names that are archetypically `non-English'.

Another extension would be to use a finer-grained distribution of
languages. The current study has three countries that are broadly
Celtic, and the influence of these languages has been discussed. However,
it would be possible to incorporate the place names of other Celtic
areas such as Breton, Cornwall and the Isle of Man. In the current
study, the place names of Breton were included in the French sample,
and the place names of Cornwall and the Isle of Man were incorporated
into the English sample. A higher level of accuracy and a subsequent
more informative score could be achieved if these language groups
were given their own representation in the ensemble of classifiers
and removed from the French and English samples. Similarly, there
are five Germanic languages included in the current study (six if
we include English), and the possibility exists to use the place names
of the Faeroe Islands, and Frisian place names to get a finer gained
analysis.

One issue that could be covered in a subsequent study is the influence
of non-Indo European languages on English place names. We have seen
that there are 475 English place names that are not classifiable as
English. Inspection of these suggests that some of them are unrelated
to the 11 languages used in this study. It would be interesting to
see how many of these closely resemble the place names of the Basque
country, Finland, Estonia and the Sami.

\hfill{}\\
\begin{flushleft}

%\section{Data Availability Statement}
\textbf {Data Availability Statement}\\
\hfill{}\\
The data used in this study are available at Figshare (https://figshare.com/), DOI: https://doi.org/10.6084/m9.figshare.24916158.v1.\\
\hfill{}\\

\textbf {Research Funding Statement}\\
\hfill{}\\
This research was supported by an Australian Government RTP Scholarship. \\
\hfill{}\\

\textbf {Disclosure Statement}\\
\hfill{}\\
The author reports there are no known competing interests to declare.\\
\end{flushleft}

\bibliographystyle{unsrt}
\bibliography{bib}

\end{document}